\documentclass[english]{lni}
\usepackage{booktabs}

\begin{document}


\title[Automated Self-Refinement and Correction]{Self-Refinement Strategies for LLM-based Product Attribute Value Extraction}


 \author[1]{Alexander Brinkmann}{alexander.brinkmann@uni-mannheim.de}{0000-0002-9379-2048}
 \author[1]{Christian Bizer}{christian.bizer@uni-mannheim.de}{0000-0003-2367-0237}
 \affil[1]{University of Mannheim\\Data \& Web Science Group\\Schloss\\68161 Mannheim\\Germany}
\maketitle

\begin{abstract}
Structured product data, in the form of attribute-value pairs, is essential for e-commerce platforms to support features such as faceted product search and attribute-based product comparison. However, vendors often provide unstructured product descriptions, making attribute value extraction necessary to ensure data consistency and usability.
Large language models (LLMs) have demonstrated their potential for product attribute value extraction in few-shot scenarios. Recent research has shown that self-refinement techniques can improve the performance of LLMs on tasks such as code generation and text-to-SQL translation. For other tasks, the application of these techniques has resulted in increased costs due to processing additional tokens, without achieving any improvement in performance.
This paper investigates applying two self-refinement techniques — error-based prompt rewriting and self-correction — to the product attribute value extraction task. The self-refinement techniques are evaluated across zero-shot, few-shot in-context learning, and fine-tuning scenarios using GPT-4o. The experiments show that both self-refinement techniques fail to significantly improve the extraction performance while substantially increasing processing costs. For scenarios with development data, fine-tuning yields the highest performance, while the ramp-up costs of fine-tuning are balanced out as the amount of product descriptions increases.


\end{abstract}
\begin{keywords}
Self-Refinement \and Information Extraction \and Large Language Models \and E-Commerce 
\end{keywords}

\section{Introduction}
\label{sec:introduction}

Large Language Models (LLMs), such as OpenAI's GPT-4o, have been successfully applied to a wide range of tasks, including information extraction tasks such as extracting product attribute values from product descriptions~\cite{brinkmann_extractgpt_2025}. In order to be effective on these tasks, LLMs often rely on few-shot in-context learning and fine-tuning. Recently, methods for the automated self-refinement of prompts~\cite{madaan_self-refine_2023} and the self-review and self-correction of model decisions~\cite{pan_automatically_2024} have emerged and are successfully applied for tasks such as code generation~\cite{madaan_self-refine_2023} and text-to-SQL translation~\cite{pourreza_din-sql_2023}.
At the same time, an increasing body of research~\cite{olausson_is_2023, huang_large_2023} criticises the self-refinement approaches as for other tasks, they do not significantly improve performance while 
increasing the processing costs heavily due to additional tokens that need to be processed. 
This paper critically evaluates two self-refinement techniques for extracting attribute values from product descriptions: error-based prompt rewriting and self-correction.
Error-based prompt rewriting improves the attribute definitions within prompts by analyzing errors made by the model on validation examples.
Self-correction reviews and updates the initial output of an LLM if it spots wrongly extracted values.
These two self-refinement techniques are chosen as they can be applied in a fully automated fashion and cover self-refinement during prompt engineering (error-based prompt rewriting) and post hoc self-refinement of the LLM's output (self-correction).
The self-refinement techniques are evaluated in zero-shot, few-shot in-context learning and fine-tuning scenarios.
This paper makes the following contributions:
\begin{itemize}
    \item The self-refinement techniques error-based prompt rewriting and self-correction are experimentally evaluated for the product attribute value extraction task. The self-refinement techniques are applied in zero-shot, few-shot in-context learning, and fine-tuning scenarios using GPT-4o.
    \item We present a detailed analysis of the impact of error-based prompt rewriting on prompt quality and the effect of self-correction on the accuracy of the extracted attribute values.
\end{itemize}

The paper is structured as follows. First, related work is reviewed. Section~\ref{sec:datasets} and Section~\ref{sec:experimental_setup} describe the datasets used and the experimental setup. Section~\ref{sec:baselines} introduces product attribute value extraction using few-shot learning and fine-tuning.
The self-refinement strategies error-based prompt rewriting and self-correction are introduced and experimentally evaluated in Section~\ref{sec:error-based_rewriting} and Section~\ref{sec:self-correction}. 
Code and data for replicating all experiments are available online\footnote{https://github.com/wbsg-uni-mannheim/SelfRefinement4ExtractGPT}.
\section{Related Work}
\label{sec:related_work}

\noindent\textbf{Attribute Value Extraction.}
Product attribute value extraction is a subtask of information extraction and focuses on extracting specific attribute values from unstructured text, such as product titles and descriptions~\cite{zhang_framework_2009,yang_mave_2022}. There are two variants of the task: closed-world, where a predefined schema specifies the target attributes, and open-world, where the set of target attributes is undefined and the extraction method needs to determine the attributes as well as their values~\cite{zhang_oa-mine_2022,xu_large_2024}. We focus on closed-world product attribute value extraction.
Early works use domain-specific rules to identify attribute values in product descriptions~\cite{zhang_framework_2009}.
Recently, many approaches have framed product attribute value extraction as a question-answering task, using the pre-trained language model (PLM) BERT to identify the target attribute value in a product description~\cite{yang_mave_2022}. 
Other related works use LLMs like GPT-4 to extract attribute values from product descriptions using different prompting techniques~\cite{fang_llm-ensemble_2024,brinkmann_extractgpt_2025}.

\noindent\textbf{Information Extraction using LLMs.}
Generative LLMs often demonstrate superior zero-shot performance compared to PLMs and exhibit higher robustness for unseen examples~\cite{brown_language_2020}. This advantage is due to extensive pre-training on large amounts of text and emergent abilities arising with large model size~\cite{wei_emergent_2022}.
LLMs have been successfully applied to information extraction tasks across various domains~\cite{xu_large_2024}. For instance, Wang et al.\cite{wang_code4struct_2023} and Parekh et al.\cite{parekh_geneva_2023} utilized OpenAI's LLMs to extract event data from unstructured text. Goel et al.~\cite{goel_llms_2023} combined LLMs with human expertise to annotate patient information in medical texts.

\noindent\textbf{Self-Refinement Techniques.}
Various techniques for model self-refinement and for correcting model outputs have been proposed recently~\cite{pan_automatically_2024}.
\cite{madaan_self-refine_2023} used automated feedback generated by an LLM to improve the readability of code. The \textit{Self-Correction} technique evaluated in this paper also relies on such automated feedback.
A critic of self-correction is that LLMs struggle to correct their responses without external feedback from tools such as code interpreters or database management systems~\cite{huang_large_2023}.
An example of how LLMs reflect on the feedback from tools to improve code successfully is given in~\cite{shinn_reflexion_2023}, while~\cite{olausson_is_2023} finds GPT-4 that is not able to generate useful feedback for fixing mistakes in code.
The \textit{Error-based Prompt Rewriting} technique that is evaluated in this paper tries to improve the initial prompt using development data as a source of external feedback.

\section{Datasets}
\label{sec:datasets}

This section introduces the benchmark datasets OA-Mine~\cite{zhang_oa-mine_2022} and AE-110k~\cite{xu_scaling_2019} that we use for the experiments. Both datasets consist of English product offers with annotated attribute-value pairs and have been used in related work~\cite{brinkmann_extractgpt_2025,yang_mave_2022}.

\noindent\textbf{OA-Mine.}
We use a subset of the human-annotated product offers of the OA-Mine dataset\footnote{https://github.com/xinyangz/OAMine/tree/main/data}~\cite{zhang_oa-mine_2022} for our experiments. The subset includes 10 product categories, with up to 80 product offers per category. Each category has between 8 to 15 attributes, resulting in a total of 115 unique attributes. Attributes with the same name in different product categories are treated as distinct attributes. We do not apply any further pre-processing to the offers. 

\noindent\textbf{AE-110K.}
The AE-110K dataset\footnote{https://raw.githubusercontent.com/lanmanok/ACL19\_Scaling\_Up\_Open\_Tagging/\linebreak master/publish\_data.txt} comprises triples of product titles, attributes and attribute values from the AliExpress Sports \& Entertainment category~\cite{xu_scaling_2019}. Product offers are derived by grouping the triples by product title. The subset includes 10 product categories, with up to 160 product offers per category. For each category, 6 to 17 attributes are known, resulting in a total of 101 unique attributes.

\noindent\textbf{Development/Test Split.} 
Table \ref{tab:dataset_statistics} contains statistics about the numbers of unique attribute-value pairs (A/V pairs), unique attribute values, and product offers for all four datasets. We use the same 60:40 development and test split based on product offers as in related work~\cite{brinkmann_extractgpt_2025}. 
The examples in the development set are labelled with ground truth attribute-value pairs.
We use the development set to sample example attribute values, to select in-context demonstrations, as a validation set for error-based prompt rewriting, and as a training set for fine-tuning.

\begin{table}[ht]
\centering
\caption{Descriptive statistics for OA-Mine and AE-110K.}
\label{tab:dataset_statistics}
\begin{tabular}{@{}l|rrr|rrr@{}}
\toprule  
        & \multicolumn{3}{l|}{OA-Mine}  & \multicolumn{3}{l}{AE-110K}   \\
        & Development & Test & Total  &  Development & Test & Total \\ \midrule
A/V Pairs   & 3,626  & 2,451 & 6,077 & 2,170  & 1,482 & 3,652 \\
Unique A/Vs & 2,400     & 1,749     &    3,637    & 587         & 454         &  854           \\
Product Offers          & 715    & 491    &  1,206   & 785   & 524 & 1,309 \\ \bottomrule             
\end{tabular}
\end{table}

\noindent\textbf{Example Extractions.} 
Table~\ref{tab:example_extractions} shows example product offer titles, target attributes and attribute values from the datasets.
The examples (a) and (b) visualize the direct extraction meaning that the extracted attribute value is a substring of the product title. 
The target output is a JSON object containing all attribute-value pairs.

\begin{table}[ht]
\centering
\caption{Example product titles and attribute-value pairs from the OA-Mine and AE-110k datasets.}
\label{tab:example_extractions}
\begin{tabular}{l|l|l}
\toprule
Dataset        & \textbf{OA-Mine}      & \textbf{AE-110k}     \\
\midrule
Category        & Vitamin& Eyewear   \\
Attributes       & Brand, Net Content,             & Sport Type, Gender              \\ 
       & Supplement Type, Dosage             & Lenses Optical, Model Number             \\ \midrule
Product Title           & NOW Supplements, Vitamin A         & Professional Men Ski Goggles Double  \\
&  (Fish Liver Oil) 25,000 IU, &  Layers UV400 Anti-fog Big Ski Mask    \\
&  Essential Nutrition, 250 Softgels      &   Skiing Snowboard Glasses     \\ \midrule
Target Output   & ´´´´json\{        &   ´´´´json\{              \\ 
& \hspace{0.25 cm} ``Brand``: ``Now Supplements``,  & \hspace{0.25 cm} ``Sport Type``: ``Skiing``, \\
& \hspace{0.25 cm} ``Net Content``: ``250``,  & \hspace{0.25 cm} ``Gender``: ``Men``,  \\
& \hspace{0.25 cm} ``Supplement Type``: ``Vitamin A``, & \hspace{0.25 cm} ``Lenses Optical``: ``UV400``,  \\
& \hspace{0.25 cm} ``Dosage``: ``25,000 IU`` &  \hspace{0.25 cm} ``Model Number``: ``n/a``\\
& \hspace{0.1 cm}\}´´´´ & \hspace{0.1 cm}\}´´´´ \\
\bottomrule
\multicolumn{1}{c}{}  & \multicolumn{1}{c}{(a)} & \multicolumn{1}{c}{(b)}  \\
\end{tabular}
\end{table}

\section{Experimental Setup}
\label{sec:experimental_setup}

We use the LLM gpt-4o-2024-08-06\footnote{https://platform.openai.com/docs/models/gp\#gpt-4o} for all experiments in this paper. GPT-4o is chosen because, at the time of experimenting for this paper, it is the LLM with the best performance on benchmarks measuring natural language understanding\footnote{https://openai.com/index/hello-gpt-4o/}.
GPT-4o is accessed via the OpenAI API. GPT-4o's temperature parameter is set to zero to reduce randomness.
We report average F1-scores of three runs for each experiment and run paired t-tests to verify with 99\% confidence if the F1-scores of two experiments are significantly different.
The F1-score is calculated by categorizing predictions into five groups as per previous works~\cite{xu_scaling_2019,yang_mave_2022,brinkmann_extractgpt_2025,brinkmann_using_2024}. The five categories are NN (no predicted value, no ground truth value), NV (predicted value, no ground truth value), VN (no predicted value, ground truth value), VC (predicted value exactly matches ground truth value), and VW (predicted value does not match ground truth value). The F1-score is derived from precision ($P = VC / (NV + VC + VW)$), recall ($R = VC/ (VN + VC + VW)$), and the formula $F1 = 2PR/(P + R)$.
Additionally, we report the average number of tokens per prompt to estimate and compare the costs of the different approaches.
It is important to mention that prompt and completion tokens are summed. OpenAI charges users different prices for input tokens (2.5\$/1M tokens) and output tokens (1.25\$/1M tokens)\footnote{OpenAI prices as of February 2025: https://openai.com/api/pricing/}.

\section{Attribute Value Extraction using In-Context Learning and Fine-Tuning}
\label{sec:baselines}
This section introduces LLM-based prompting techniques for attribute value extraction. It covers zero-shot prompts, prompts with attribute definitions generated from development data, few-shot in-context learning prompts, few-shot in-context learning prompts combined with self-consistency and fine-tuning on the development sets.

\begin{figure}[ht]
\centering
\includegraphics[width=.95\textwidth]{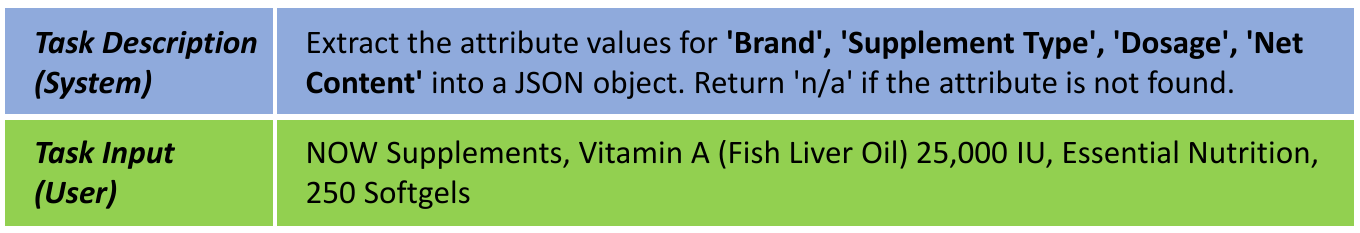}
\caption{Zero-shot prompt.}
\label{fig:zero_shot_prompts}
\end{figure}

\noindent\textbf{Zero-Shot.}
Figure~\ref{fig:zero_shot_prompts} shows the zero-shot prompt consisting of a task description and a task input. The task description is a system chat message that describes the attribute value extraction task, lists the target attributes, defines the output as a JSON object, and explains that not available attribute values should be marked with \texttt{'n\slash a'} in the output JSON object. The task input is a user chat message that contains the product description from which the attribute values are extracted. Examples of output JSON objects are shown in Table~\ref{tab:example_extractions}.

Row \textit{Zero-Shot} in Table~\ref{tab:attribute_value_extraction_results} shows the results for zero-shot attribute value extraction experiments. The F1-scores remain below 70\%, which is unsatisfying for real-world deployment. The standard deviation of the reported average F1-scores is 0.11 and 0.45 for OA-Mine and AE-110k, showing that the extracted attribute values only marginally deviate across the runs.
We use the results of the zero-shot experiment as a reference point for the upcoming experiments. We report the delta between each experimental F1-score and the zero-shot F1-score as $\Delta$ ZS. We report the ratio of the number of tokens to the number of tokens used by the zero-shot prompt as token factor (TF).

\begin{table}[ht]
\centering
\caption{Experimental results for attribute value extraction using in-context learning and fine-tuning.}
\label{tab:attribute_value_extraction_results}
\begin{tabular}{@{}l|rrrr|rrrr@{}}
\toprule
                                 & \multicolumn{4}{l|}{\textbf{OA-Mine}}                                                                                                       & \multicolumn{4}{l}{\textbf{AE-110k}}                                                                                                       \\ \midrule
                                 & \textbf{F1}                 & \textbf{$\Delta$ ZS} & \textbf{Tokens} & \textbf{TF} & \textbf{F1}                 & \textbf{$\Delta$ ZS} & \textbf{Tokens} & \textbf{TF} \\ \midrule
Zero-shot                        & 68.8 & 0.0                                             & 181                        & 1.0                     & 63.6 & 0.0                                             & 196                        & 1.0                     \\ \midrule
Attribute Definitions           & 72.2                   & +3.4                                             & 601                        & 3.3                     & 76.3                   & +12.7                                            & 577                        & 2.9                     \\ \midrule
Few-Shot                                                   & 78.6                   & +9.8                                             & 1,315                       & 7.2                     & 83.9                   & +20.3                                            & 1,351                       & 6.9                     \\
+ Self-Consistency                         & 79.3                   & +10.5                                             & 3,945                       & 21.7                    & 84.1                   & +20.4                                            & 4,053                       & 20.6                    \\
+ Attribute Definitions                   & 79.3                   & +10.5                                            & 1,760                       & 9.7                     & 85.3                   & +21.7                                            & 1,727                       & 8.8                     \\
 \midrule
Fine-Tuning                                                & 83.2                   & +14.3                                            & 172                        & 0.9                     & 85.1                   & +21.4                                            & 177                        & 0.9                     \\ \bottomrule
\end{tabular}
\end{table}

\begin{figure}[ht]
\centering
\includegraphics[width=.95\textwidth]{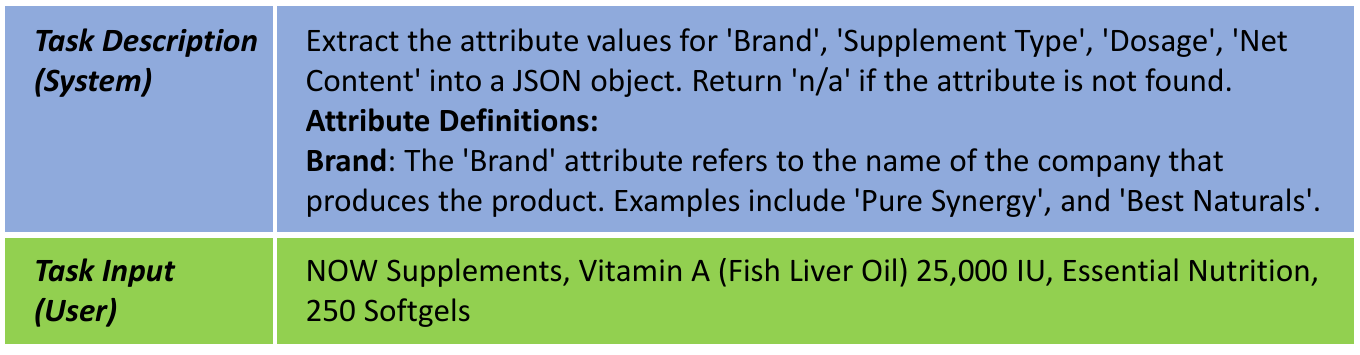}
\caption{Prompt with attribute definitions.}
\label{fig:attribute_definitions_prompts}
\end{figure}

\noindent\textbf{Attribute Definitions.}
In the attribute definitions scenario, attribute definitions are appended to the task description of the zero-shot prompt. This prompt builds on the findings of previous research demonstrating that definitions enhance the performance of LLMs for classification~\cite{peskine_definitions_2023} and extraction tasks~\cite{brinkmann_extractgpt_2025}. The example prompt in Figure~\ref{fig:attribute_definitions_prompts} depicts this extension. For demonstration purposes, it contains only the definition for the attribute \texttt{'Brand'}. In the experiments, a definition is appended for each attribute. Since the datasets OA-Mine and AE-110k do not contain attribute definitions, the definitions are generated by an LLM based on five attribute values. The attribute values are randomly sampled from the development set. Generating attribute definitions requires an average of 84 and 74 tokens per unique attribute for OA-Mine and AE-110k, respectively. 
The generated attribute definitions have an average length of 36 tokens for OA-Mine and 35 tokens for AE-110k.

Table~\ref{tab:attribute_value_extraction_results} shows that compared to the zero-shot results, the attribute definitions improve GPT-4o's F1-score by 3\% for OA-Mine and nearly 13\% for AE-110k, which are significant differences (verified using paired t-tests with 99\% confidence), but the prompts including the definitions are approximately 3 times as long as the zero-shot prompts (TF=3.3 and TF=2.9). This result underlines the usefulness of attribute definitions for extracting product attribute values.

\begin{figure}[ht]
\centering
\includegraphics[width=.95\textwidth]{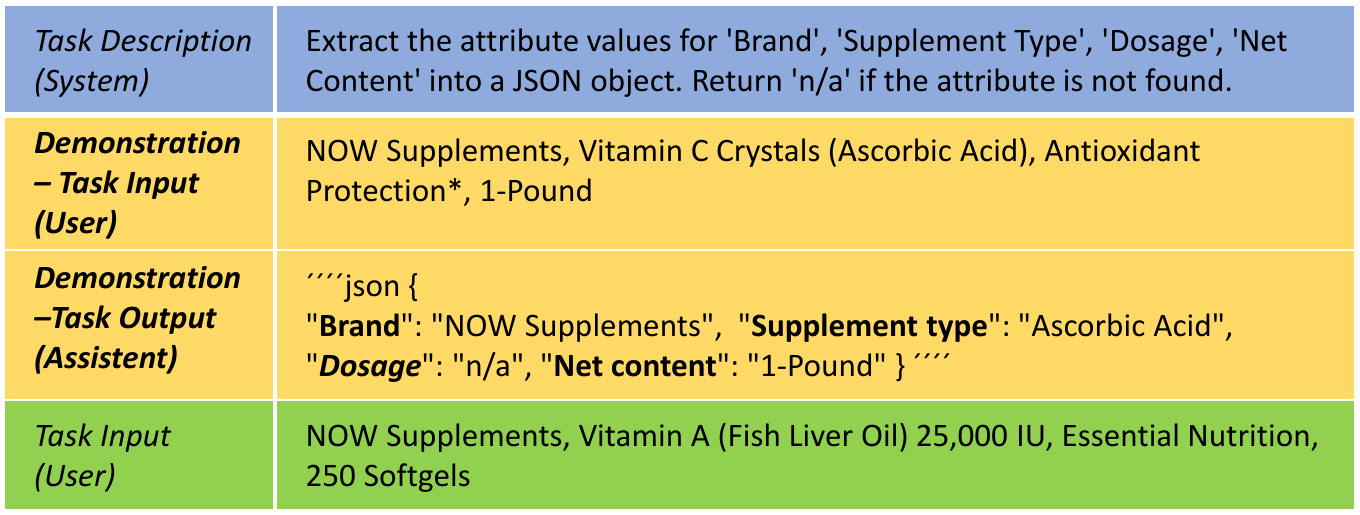}
\caption{Few-shot prompt.}
\label{fig:few-shot_prompt}
\end{figure}

\noindent\textbf{Few-Shot In-Context Learning.}
In the few-shot scenario, demonstrations with annotated attribute-value pairs are presumably available as a development set.
For few-shot learning, the zero-shot prompt and the prompt with attribute definitions are extended with demonstrations. Figure~\ref{fig:few-shot_prompt} shows the extension of the zero-shot prompt. Each demonstration consists of a demonstration task input and a demonstration task output. The demonstration task input is a user message containing a product description. The demonstration task output is an assistant message with the extracted attribute-value pairs formatted as a JSON object.
We use demonstrations that are semantically similar to the current product offer. For selecting these demonstrations, the demonstrations of the development set are embedded using OpenAI's embedding model \texttt{text-embedding-ada-002}\footnote{https://platform.openai.com/docs/guides/embeddings/}. The embedded demonstrations with the greatest cosine similarity to the embedded task input are considered to be semantically similar. Following~\cite{brinkmann_extractgpt_2025}, we add 10 demonstrations to each prompt. 

Row \textit{Few-Shot} in Table~\ref{tab:attribute_value_extraction_results} reports the results of the few-shot experiments.
As expected, few-shot in-context learning with semantically similar demonstrations significantly improves GPT-4o's F1-score but more than doubles the number of tokens per prompt compared to adding attribute definitions (e.g. the TF for OA-Mine rises from 3.3 to 7.2). The gains show that GPT-4o requires demonstrations to achieve F1-scores above or at least close to 80\%. Attribute definitions increase the F1-score by 1\%, which is a significant increase according to the t-tests, but adds around 400 tokens to the prompt (TF=9.7 and TF=8.8). 

\noindent\textbf{Few-Shot Self-Consistency.}
Self-consistency extends the few-shot in-context learning prompt by sampling three outputs from the LLM and selecting the most consistent answer through majority voting for each attribute.
Each run attribute order in the prompt is shuffled to avoid an ordering bias~\cite{zhang_survey_2025}.
Related work proposes self-consistency as an alternative to self-refinement with similar token usage~\cite{huang_large_2023}. 
Row \textit{Self-consistency} in Table~\ref{tab:attribute_value_extraction_results} reports the experimental results.
Due to the three runs, few-shot in-context learning with self-consistency costs three times the number of tokens compared to few-shot in-context learning. The ensemble runs have a low deviation in the extracted attribute values explaining the insignificant performance changes.

\noindent\textbf{Fine-Tuning.}
In-context learning provides task-specific knowledge to the LLM via the prompt.
Fine-tuning uses training data to encode task-specific knowledge into the parameters of the LLM.
At runtime, this task-specific knowledge is implicitly used to extract attribute values.
In preparation for fine-tuning, the development records are formatted with the zero-shot prompt consisting of a task description, an input and an output containing attribute-value pairs. 
To execute the fine-tuning, the pre-processed datasets are uploaded to OpenAI's fine-tuning API\footnote{https://platform.openai.com/docs/guides/fine-tuning} and GPT-4o is trained for three epochs on the uploaded datasets using OpenAI's default parameters.
Row \textit{Fine-Tuning} in Table~\ref{tab:attribute_value_extraction_results} reports the fine-tuning results for GPT-4o.
The fine-tuned GPT-4o LLMs achieve the highest average F1-score with the lowest token usage during application (TF=0.9).
But especially on the dataset AE-110k, few-shot in-context learning is a competitive alternative to fine-tuning.
The initial investment required for fine-tuning is 394k and 431k tokens, respectively, for OA-Mine and AE-110k.
Based on the number of tokens and the token cost per product offer, a break-even point concerning the number of test product offers for using fine-tuning GPT-4o instead of few-shot in-context learning can be calculated as follows: $FineTuning Cost/(Token Cost_{FewShot} - Token Cost_{FineTuning}$). Using this formula the average break-even point for the datasets is 6,666 product offers. For scenarios in which attribute values have to be extracted from more than 6,666 product offers, it is cheaper to invest in fine-tuning than to use the base model with the longer in-context learning prompts.

\section{Error-based Prompt Rewriting}
\label{sec:error-based_rewriting}
This section introduces and evaluates the self-refinement technique of error-based prompt rewriting. The method is combined with the zero-shot and few-shot in-context learning prompts that use attribute definitions introduced in Section~\ref{sec:baselines}.

\begin{figure}[ht]
\centering
\includegraphics[width=.95\textwidth]{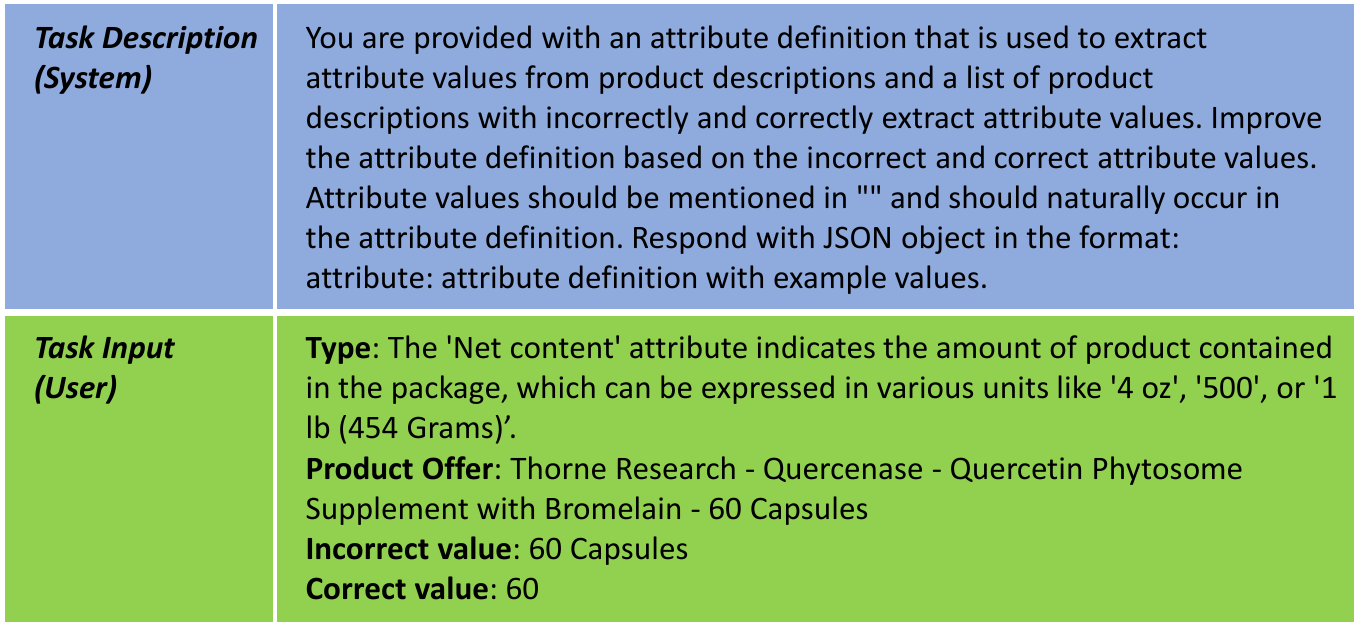}
\caption{Prompt for error-based rewriting of attribute definitions.}
\label{fig:error_based_prompt_rewriting}
\end{figure}

\noindent\textbf{Prompting Technique.}
Error-based prompt rewriting uses the development set to improve the attribute definitions in the prompts. It assumes that better attribute definitions improve the product attribute value extraction~\cite{peskine_definitions_2023}. For error-based prompt rewriting, the prompts with attribute definitions are run on five randomly selected product offers from the development set for each category to extract attribute values. 
The extracted attribute values are compared to the ground truth from the development set to identify incorrectly extracted values. For each attribute with extraction errors, the prompt shown in Figure~\ref{fig:error_based_prompt_rewriting} is populated with the existing attribute definition and a list of product descriptions with incorrectly and correctly extracted attribute values. 
In the depicted example, '60' instead of '60 Capsules' is expected. The LLM responds to the prompt with a rewritten attribute definition. Table~\ref{tab:rewritten_attribute_definitions} illustrates how the attribute definition for the \texttt{Net Content} is rewritten to be more specific. Error-based prompt rewriting is repeated three times to evaluate the product attribute value extraction on up to 15 product descriptions to improve the attribute definitions.

\begin{table}[]
\centering
\caption{Definitions for the attribute \texttt{'Net Content'} of a vitamin product.}
\label{tab:rewritten_attribute_definitions}
\begin{tabular}{@{}l|l|l@{}}
\toprule
\textbf{Extraction Error} & \textbf{Attribute Definition}   & \textbf{Rewritten Attribute Definition} \\ \midrule
Extracted Value: & The 'Net Content' attribute & The 'Net Content' attribute specifies the \\
"60 Capsules" & indicates the amount of & quantity of product within a package. For\\ 
 & product contained in  & products like capsules, or softgels, it is \\
 Expected Value: & the package, which can be &  expressed as a numerical value, such as '60'. \\
 "60" & expressed in various units  & For liquid products, it is expressed in fluid \\
 & like '4 oz', '500', or & ounces, such as '0.08 Fl Oz'. For products \\
 & '1 lb (454 Grams)'. &  sold by weight, it should be expressed in \\
 & & pounds or grams, such as '1 lb (454 Grams)'. \\ \bottomrule
\end{tabular}
\end{table}

\noindent\textbf{Discussion of Results.}
Rewriting attribute definitions increases token usage. In the zero-shot scenario, 270k tokens are required on OA-Mine for rewriting attribute definitions, while 350k tokens are consumed on AE-110k. These values rise to 378k and 389k tokens in the few-shot scenario.
Table~\ref{tab:error_based_rewriting_results} reports the experimental results of the product attribute extraction without and with rewritten attribute definitions. In the zero-shot scenario, the rewritten definitions insignificantly degrade GPT-4o's performance (verified by t-tests with 99\% confidence) and cause an increased token usage (e.g. the TF for OA-Mine rises from 3.3 to 8.6). In the few-shot learning configuration, the performance does not significantly change, though token usage rises again (e.g. the TF for OA-Mine rises from 9.7 to 13.9).

\begin{table}[]
\centering
\caption{Experimental results for error-based prompt rewriting.}
\label{tab:error_based_rewriting_results}
\begin{tabular}{@{}l|rrrr|rrrr@{}}
\toprule
                                 & \multicolumn{4}{l|}{\textbf{OA-Mine}}                                                                                                       & \multicolumn{4}{l}{\textbf{AE-110k}}                                                                                                       \\ \midrule
                                 & \textbf{F1}                 & \textbf{$\Delta$ ZS} & \textbf{Tokens} & \textbf{TF} & \textbf{F1}                 & \textbf{$\Delta$ ZS} & \textbf{Tokens} & \textbf{TF} \\ \midrule
Attribute Definitions           & 72.2                   & +3.4                                             & 601                        & 3.3                     & 76.3                   & +12.7                                            & 577                        & 2.9                     \\
+ Rewriting      & 71.8                   & +2.9                                             & 1,569                       & 8.6                     & 74.4                   & +10.8                                            & 1,872                       & 9.5                     \\ \midrule
Few Shot + Attr. Defs.                   & 79.3                   & +10.5                                            & 1,760                       & 9.7                     & 85.3                   & +21.7                                            & 1,727                       & 8.8                     \\
+ Rewriting & 79.0 & +10.2 & 2,520 & 13.9 & 85.5 & +21.8 & 2,650 & 13.5 \\
\bottomrule
\end{tabular}
\end{table}

\noindent\textbf{Analysis of Rewritten Attribute Definitions.}
The analysis of error-based attribute definition rewriting focuses on the OA-Mine dataset in the zero-shot scenario. To reduce the manual effort of analyzing all changes, GPT-4o is used to analyze the attribute definition. GPT-4o analyzes the length of the attribute definitions, the level of detail of the original and the rewritten attribute definition, counts the number of example values per attribute definition and checks if values are excluded in an attribute definition. We manually review 10 analyzed rewriting operations to estimate how well GPT-4o performs this analysis. In the manual assessment, the level of detail and the mention of excluded example values are always correct. The example value counts are in 80\% of the cases correct. With a tolerance range of two, all example value counts are valid. Hence, GPT-4o is useful for this analysis. Over three iterations, GPT-4o completed 176 out of 345 possible rewrites. The 345 possible rewrites are calculated by multiplying the number of unique attributes (115) in OA-mine by three because every iteration each unique attribute definition can but must not be rewritten once. The quantitative analysis of the attribute definitions shows that 97\% of the rewritten attribute definitions are longer than the original attribute definition. The longer attribute definitions explain the higher TF reported in Table~\ref{tab:error_based_rewriting_results}. 93\% of GPT-4o's rewriting operations enhance the level of detail. In 65\% of the operations, the level of detail is increased by adding example values and in 31\% of the operations, the level of detail is increased by explicitly excluding attribute values. The enhanced level of detail through additional information and the inclusion and exclusion of attribute values leads to an overfitting of the attribute definitions to the development set, which harms the LLM's performance on the test set.

\section{Self-Correction}
\label{sec:self-correction}
This section introduces and evaluates the post hoc self-refinement technique self-correction. Self-correction can be combined with all prompts introduced in Section~\ref{sec:baselines}.

\begin{figure}[ht]
\centering
\includegraphics[width=.95\textwidth]{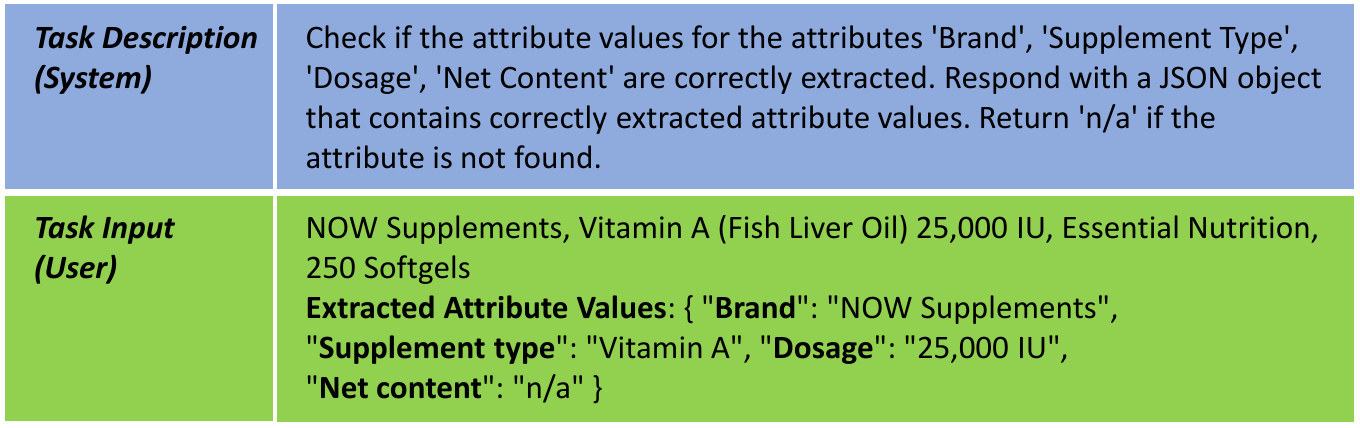}
\caption{Prompt for self-correction.}
\label{fig:prompt_self-correction}
\end{figure}

\noindent\textbf{Prompting Technique.}
Motivated by related work~\cite{madaan_self-refine_2023}, the LLM post hoc reviews and updates its initially extracted attribute values. Therefore, a first prompt instructs the LLM to extract attribute-value pairs from the input. This first prompt is one of the prompts introduced in Section~\ref{sec:baselines} depending on the scenario (zero-shot, few-shot in-context learning or fine-tuning). The output of the first prompt is sent to the same LLM again with a request to reflect on and correct erroneously extracted attribute values. Figure~\ref{fig:prompt_self-correction} illustrates the second self-correction prompt. A criticism of related work on the original self-refinement paper is that the refinement prompt provides additional task-related information~\cite{huang_large_2023}. To ensure that similar information is provided by the initial prompt and the self-correction prompt, attribute definitions and few-shot in-context learning demonstrations are added to the prompt if the initial prompt contains them. In the fine-tuning scenario, the fine-tuned GPT-4o executes the self-correction prompt.

\noindent\textbf{Discussion of Results.}
Table~\ref{tab:zero-shot_results} shows that in the zero-shot scenario, the performance gains from self-correction are insignificant (verified by t-tests with 99\% confidence) and do not justify the substantial increase in computational cost (TF=2.6). In the few-shot scenario with in-context learning demonstrations, self-correction does not significantly alter GPT-4o's performance while more than doubling token usage (e.g. the TF for OA-Mine rises from 7.2 to 15.2). 
In the fine-tuning scenario, the performance differences are insignificant but again the token usage doubles (e.g. the TF for OA-Mine rises from 0.9 to 2.5).
These findings indicate that self-correction is not cost-effective and only marginally influences the performance of GPT-4o for attribute value extraction across different learning scenarios.

\begin{table}[]
\centering
\caption{Experimental results for self-correction.}
\label{tab:zero-shot_results}
\begin{tabular}{@{}l|rrrr|rrrr@{}}
\toprule
 & \multicolumn{4}{l|}{\textbf{OA-Mine}}       & \multicolumn{4}{l}{\textbf{AE-110k}}       \\ \midrule
 & \textbf{F1} & \textbf{$\Delta$ ZS} & \textbf{Tokens} & \textbf{TF} & \textbf{F1} & \textbf{$\Delta$ ZS} & \textbf{Tokens} & \textbf{TF} \\ \midrule
Zero-shot        & 68.8 & 0.0             & 181        & 1.0     & 63.6 & 0.0             & 196        & 1.0     \\
 + Self-Correction & 69.3 & +0.4             & 467        & 2.6     & 63.8 & +0.1             & 503        & 2.6     \\ \midrule
Attribute Definitions           & 72.2   & +3.4             & 601        & 3.3     & 76.3   & +12.7            & 577        & 2.9     \\
+ Self-Correction & 72.5   & +3.6             & 1,305       & 7.2     & 77.2   & +13.5            & 1,285       & 6.5     \\ \midrule
Few-Shot   & 78.6   & +9.8             & 1,315       & 7.2     & 83.9   & +20.3            & 1,351       & 6.9     \\
+ Self-Correction          & 78.5   & +9.6             & 2,751       & 15.2    & 83.7   & +20.0            & 2,815       & 14.3    \\
+ Attribute Definitions   & 79.3   & +10.5            & 1,760       & 9.7     & 85.3   & +21.7            & 1,727       & 8.8     \\

 + Attr. Def. \& Self-Corr. & 78.7   & +9.8             & 3,589       & 19.8    & 84.9   & +21.3            & 3,585       & 18.3    \\\midrule
Fine-Tuning& 83.2   & +14.3            & 172        & 0.9     & 85.1   & +21.4            & 177        & 0.9     \\ 
+ Self-Correction          & 82.9	& +14.1 &	449 &	2.5 &	85.2 &	+21.6 &	457 &	2.3\\
\bottomrule
\end{tabular}
\end{table}

\noindent\textbf{Analysis of Self-Corrected Attribute Values.}
We analyze the self-correction in detail by distinguishing three possible outcomes of the self-correction step: (1) a wrong value is corrected (Improvement), (2) a previously correct value is corrupted (Corruption), and (3) the update of the extracted value does not correct a wrong value but just changes it (Still wrong). Table~\ref{tab:self-correction_scenarios} provides examples of the three outcomes. The target attribute value is underlined in the product description. A quantitative analysis of the zero-shot scenario shows that 64\% of the 165 attribute value updates on OA-Mine and 90\% of the 423 attribute value updates have no impact because the extracted value and the updated value are incorrect.
Across the zero-shot, few-shot in-context learning and fine-tuning scenarios, corruptions happen more often than improvements explaining the marginal decrease in performance when self-correction is applied. For instance, in the zero-shot scenario on OA-Mine 19\% of the corrections are improvements while 16\% of the corrections are corruptions.
On both datasets, the amount of corrected values decreases from zero-shot to fine-tuning. For example on OA-Mine, 165 values are updated in the zero-shot scenario while 106 values are updated after fine-tuning.
The LLM seems to correct attribute values where it is undecided between two possible values.
These undecided attribute values change with in-context learning demonstrations.
The updated extracted attribute values of the zero-shot and the few-shot in-context learning scenario overlap only by 13 and 25 values for OA-Mine and AE-110k, respectively.

\begin{table}[]
\centering
\caption{Possible outcomes of self-correction.}
\label{tab:self-correction_scenarios}
\begin{tabular}{@{}l|l|l|l|l@{}}
\toprule
\textbf{Outcome}   & \textbf{Attribute}   & \textbf{Product Description} & \textbf{Extracted Value} & \textbf{Corrected Value} \\ \midrule
Improvement    & Net Content & \begin{tabular}[c]{@{}l@{}}Nature's Path Organic \\ Oatmeal, (Pack of 6, \\ \underline{11.3 Oz} Boxes)\end{tabular} & 11.3 Oz Boxes \textcolor{red}{$\times$}  & 11.3 Oz \textcolor{green}{\checkmark}  \\ \midrule
Corruption  & Pack size   & \begin{tabular}[c]{@{}l@{}}Goddess Garden - \\ SPF 50 Sunscreen \\ Stick - \underline{1 Unit}\end{tabular}     & 1 Unit \textcolor{green}{\checkmark}         & n/a  \textcolor{red}{$\times$}           \\ \midrule
Still wrong & Item form   & \begin{tabular}[c]{@{}l@{}}Good Natured Lavender\\ \underline{Laundry Soda/Detergent} \\ 52 load bag 32 oz.\end{tabular}       & Soda/Detergent \textcolor{red}{$\times$}  & Laundry Soda  \textcolor{red}{$\times$}  \\ \bottomrule
\end{tabular}
\end{table}

\section{Conclusion}
\label{sec:conclusion}

This paper evaluated self-refinement strategies for large language models (LLMs) in the context of product attribute value extraction.
The experimental evaluation examined two self-refinement techniques: error-based prompt rewriting and self-correction, across zero-shot, few-shot in-context learning, and fine-tuning scenarios using GPT-4o. Error-based prompt rewriting and self-correction increased computational cost due to higher token consumption without significant gains in extraction performance.
While error-based prompt rewriting improved the level of detail of attribute definitions by adding and excluding example values, this likely led to overfitting to the development set.
Self-correction occasionally corrected wrongly extracted attribute values but also introduced new errors, leading to insignificant changes in the extraction performance.
Overall, fine-tuning without self-refinement achieved the highest F1-score and is the most cost-efficient approach for scenarios where attribute values need to be extracted from a large number of product descriptions.

\bibliography{references}
\end{document}